\documentclass{article} 
\usepackage{iclr2027_conference,times}

\usepackage[T1]{fontenc}
\usepackage[utf8]{inputenc}

\usepackage{amsmath,amssymb,amsfonts}
\usepackage{bm}

\usepackage{booktabs}
\usepackage{array}
\usepackage{multirow}
\usepackage[table]{xcolor}

\usepackage{url}
\usepackage{graphicx}
\usepackage{float}
\usepackage[hidelinks]{hyperref}
\hypersetup{
  pdftitle={RACE: Relation-Level Counterfactual Explanations for Heterogeneous Graph Neural Networks},
  pdfauthor={Yuxiang Yao, Zijun Zhao}
}

\newcommand{\method}{\textsc{Race}}
\newcommand{\G}{\mathcal{G}}
\newcommand{\Vset}{\mathcal{V}}

\title{RACE: Relation-Level Counterfactual Explanations for Heterogeneous Graph Neural Networks}

\author{Yuxiang Yao$^1$\thanks{Equal contribution.} \qquad
Zijun Zhao$^2$\footnotemark[1] \thanks{Corresponding author.}\\
$^1$Project Management Department, Research Center, China Life Insurance Company Ltd.\\
$^2$School of Computer Science and Technology, Beijing Institute of Technology\\
\texttt{yaoyuxiangyyx2023@e-chinalife.com} \qquad \texttt{zhaozijun@bit.edu.cn}
}

\iclrfinalcopy 

\begin{document}

\maketitle
\lhead{Preprint} 

\begin{abstract}
Counterfactual explanations of graph neural networks identify edge deletions that flip a prediction. On heterogeneous graphs, however, existing methods first collapse the graph into untyped edges, so they cannot answer the question a domain expert actually asks: \emph{which relation type} drives this prediction? We present \method{} (\textbf{R}elation-\textbf{A}ware \textbf{C}ounterfactual \textbf{E}xplanations), which gives this question an exact, per-instance answer. For every explained instance, an exhaustive search over relation subsets returns the certified minimum relation-deletion set that flips the prediction---or an explicit report that no such deletion exists; each relation-level answer is then refined into a typed edge set within the attributed relations, verified on the discrete model by single-edge restoration. The relation-level answer is exact and deterministic given the frozen backbone, whereas soft-mask baselines vary by $6$--$8$~pp in success rate across runs differing only in random ordering. On ACM, a Cora-derived graph, and ogbn-mag, \method{} improves counterfactual success rate over the strongest baseline by up to $+2.7$~pp while deleting fewer edges, and attains the highest success rate among all same-task baselines on every dataset; the advantage reproduces across four backbones on ogbn-arXiv and on DBLP, with cross-seed relation-set agreement up to $0.89$. A synthetic study with known generating mechanisms confirms that the search recovers the relation the trained model actually relies on---and reports infeasibility rather than fabricating an attribution when the model has learned none---so the explanations stay trustworthy exactly where explanations matter.
\end{abstract}

\section{Introduction}
\label{sec:intro}

Graph neural networks (GNNs) are the workhorse for learning on graph-structured data, yet their predictions remain opaque. \emph{Counterfactual} explanations address opacity by identifying the minimal input perturbation that \emph{flips} the prediction: CF-GNNExplainer~\cite{lucic2022cfgnnexplainer}, CF\textsuperscript{2}~\cite{tan2022cf2}, and RCExplainer~\cite{bajaj2021rcexplainer} delete edges until the prediction changes, and recent work keeps extending this line~\cite{zhang2026atexcf}. Real-world graphs, however, are rarely homogeneous: academic networks, knowledge graphs, and e-commerce platforms are heterogeneous information networks in which co-authorship, citation, and shared-subject links coexist~\cite{shi2016survey}. On such data the existing methods have two defects. First, they \emph{collapse} the heterogeneous graph into a pairwise graph, discarding relation-type semantics, so the resulting explanation cannot answer ``\emph{which relation type} drives the prediction''---the question a domain expert asks of a heterogeneous model. Second, even in their native regime, their edge-level outputs are a scattered set of individual edges whose semantics are not aggregable into a human-readable statement such as ``co-authorship is what this prediction relies on''.

We close this gap with \method{}, a framework that (i)~preserves relation-type semantics in the explained model, (ii)~computes \emph{instance-level} minimal relation-deletion sets by exact search, (iii)~refines them into \emph{verified}, single-edge-restoration-irreducible edge deletions inside the attributed relations (a relation-to-edge hierarchy), and (iv)~evaluates everything with deletion- and sufficiency-fidelity metrics plus explicit coverage, cost, and stability reporting. We make three contributions:

\begin{itemize}
\item \textbf{Instance-level relation counterfactuals by exact search.} For each explained instance $v$ we solve $S_v^{*} = \arg\min_{S} C_v(S)$ subject to the prediction flipping (a margin constraint), with a lexicographic cost (number of relations, then local edge fraction, then margin). Because $K$ is small, enumeration over the $2^K$ subsets is exact, deterministic, and cheap (${\le}10$~s at $K{=}10$); every instance receives either this certified answer or an explicit infeasibility report---no force-attribution---and the resulting coverage is disclosed. A budgeted branch-and-bound search extends the same contract to larger $K$ (RQ6). We further document \emph{non-monotonicity}: some predictions flip under a relation subset but not under full removal, evidence that aggregate flip-rate searches cannot substitute for the instance-level problem.
\item \textbf{A verified hierarchical relation-to-edge pipeline.} The edge-level stage deletes the attributed relations and then \emph{verifies} the intervention on the discrete model by restoring edges one at a time along the saliency order while the flip persists, repeated to a fixpoint---so the reported set carries a single-edge-restoration irreducibility certificate (or an explicit budget-limited status), evaluated on the instance's exact local computation subgraph. Under a strict same-frozen-typed-backbone, instance-targeted protocol, the hierarchical explainer beats the strongest baseline (CF\textsuperscript{2}-hetero) on both success rate ($+0.8$/$+2.7$/$+1.4$~pp on ACM/Cora/ogbn-mag, $p=0.002$ raw, Holm-corrected ${\le}0.062$, ten seeds) and edge cost ($-1.9$/$-1.0$/$-2.7$~pp), reproduces on ogbn-arXiv across four backbones and on DBLP, yields cross-seed relation-set agreement of $0.78$--$0.89$ on the separable datasets, and attains the highest success rate among the same-task counterfactual baselines under the same frozen checkpoint.
\item \textbf{A fidelity framework and an identifiability characterization.} We adopt the deletion/sufficiency statistics as descriptive fidelity metrics---Necessity Fidelity (NF) and Sufficiency Fidelity (SF)---reserving causal language for settings with an explicit generating mechanism; we characterize \emph{when} relation-level attribution is identifiable (causally separable relations), and---via a five-mechanism synthetic study separating model-dependence from data-causal recovery---show that the search recovers the relation the trained model actually relies on and reports infeasibility instead of fabricating an attribution when the model has not learned a mechanism.
\end{itemize}

\section{Related Work}
\label{sec:related}

\textbf{Explaining GNNs.} GNNExplainer~\cite{ying2019gnnexplainer} learns a mask over edges and features that \emph{preserves} the prediction, and PGExplainer~\cite{luo2020pgexplainer} amortizes this into a generator; both are factual and cannot answer counterfactual queries. Gradient attributions---Saliency~\cite{baldassarre2019explainability}, GradCAM~\cite{pope2019explainability}, Integrated Gradients~\cite{sundararajan2017axiomatic}---and SubgraphX~\cite{yuan2021subgraphx} score inputs without a flip guarantee. None provides relation-type granularity.

\textbf{Counterfactual explanations on graphs.} CF-GNNExplainer~\cite{lucic2022cfgnnexplainer} formalizes the counterfactual explanation as a minimal set of edge deletions that flips the prediction; CF\textsuperscript{2}~\cite{tan2022cf2} adds factual reasoning and a validity term; RCExplainer~\cite{bajaj2021rcexplainer} uses a hinge objective; MEG~\cite{numeroso2021meg} searches binary masks with a genetic algorithm; NSEG~\cite{cai2025pns} optimizes a necessity-plus-sufficiency lower bound; InduCE~\cite{verma2024induce} amortizes the counterfactual objective into an inductive generator; C2Explainer~\cite{ma2025c2explainer} makes the counterfactual customizable through prior constraints. All operate on homogeneous pairwise graphs: none attributes an explanation to a \emph{relation type}, because the graph they consume no longer contains types. What this line lacks is not post-hoc evaluation---prior methods do evaluate their final counterfactual on the discrete model when reporting validity---but a \emph{selection} procedure that consults the discrete model at every step, per instance, and reports infeasibility instead of force-attributing---the ingredients \method{} contributes.

\textbf{Probabilities of causation and heterogeneous GNNs.} Pearl introduced necessity, sufficiency, and their conjunction as counterfactual interpretations of causation~\cite{pearl1999probabilities,tian2000probabilities}; \cite{cai2025pns} brought this framework to graph explanations. We adopt their evaluation conventions as descriptive fidelity metrics (NF/SF) and reserve causal language for synthetic data with an explicit generating mechanism. HAN~\cite{wang2019heterogeneous}, HGT~\cite{hu2020heterogeneous}, and R-GCN~\cite{schlichtkrull2018rgcn} organize heterogeneous message passing; type-aware attribution methods remain factual and post-hoc. Heterogeneous explainability has not been combined with (i)~minimal \emph{instance-level} counterfactual interventions, (ii)~relation-type-level intervention units with explicit coverage, and (iii)~per-instance verification on the discrete model---the three ingredients \method{} contributes.

\section{Methodology}
\label{sec:method}

\subsection{Preliminaries and Problem Formulation}
\label{sec:prelim}

A heterogeneous graph is a tuple $\G = (\Vset, \{E_r\}_{r=1}^{K}, X)$ with relation edge sets $E_r \subseteq \Vset \times \Vset$, $K$ relation types ($K \le 10$ in our benchmarks), and node features $X \in \mathbb{R}^{N \times d}$. A frozen graph neural classifier $f$ maps $\G$ to class probabilities $f(\G) \in [0,1]^{N \times C}$. For an instance $v$ with prediction $\hat y_v = \arg\max_c f(\G)_{v,c}$, we write the margin
\begin{equation}
M_v(\G) \;=\; f(\G)_{v,\hat y_v} - \max_{c \neq \hat y_v} f(\G)_{v,c},
\label{eq:margin}
\end{equation}
so a flip means $M_v \le -\kappa$ for a target confidence $\kappa \ge 0$. We study two intervention granularities: \emph{relation-type-level} (delete every edge of one or more relations, written $\G \setminus S$ for $S \subseteq \{1,\dots,K\}$) and \emph{edge-level} (delete a typed edge subset, written $\G \odot m$ for per-relation keep-masks $m^r \in [0,1]^{|E_r|}$). Throughout, $N_L(v)$ denotes the backward ball of radius $L$ around $v$ (the receptive field of an $L$-layer backbone), and $E_L(v)$ the edges of the $L$-layer computation subgraph of $v$.

\subsection{Relation-Aware Heterogeneous Backbone}
\label{sec:backbone}

The explained model preserves relation semantics through a simplified heterogeneous graph transformer. For each relation $r$, source features are projected by a dedicated matrix and edges are scored by additive attention:
\begin{align}
h^{r}_{v} &= W_r x_v, \nonumber\\
e^{r}_{uv} &= \mathrm{LeakyReLU}\!\left(a_r^{\top} \big[ W_r x_u \,\Vert\, W_r x_v \big]\right), \nonumber\\
\alpha^{r}_{uv} &= \mathrm{softmax}_{v \in \mathcal{N}_r(u)}\!\left(e^{r}_{uv}\right),
\label{eq:attn}
\end{align}
and the layer update combines relations with learnable relation weights $\lambda_r$:
\begin{equation}
x'_u = \mathrm{ReLU}\!\left(\mathrm{LayerNorm}\!\left(W_{\mathrm{self}}\, x_u
+ \sum_{r=1}^{K} \lambda_r \sum_{v \in \mathcal{N}_r(u)} \alpha^{r}_{uv}\, W_r x_v\right)\right).
\label{eq:update}
\end{equation}
Stacking $L$ such layers and a linear head yields $f$. On ACM this backbone attains $0.560\pm0.007$ accuracy, $+3.4$~pp over a collapsed pairwise backbone trained identically; HAN, HGT, and the \method{} backbone tie at $0.56$--$0.57$ under the shared protocol, while residual-free GCN/GAT attain $0.23$---the accuracy edge is a property of the typed residual design.

\subsection{Edge-Level Verified Counterfactual}
\label{sec:edge}

\textbf{Per-relation typed masks.} For each relation $r$ we learn a keep-mask $m^{r} \in [0,1]^{|E_r|}$ (Gumbel-Sigmoid relaxation~\cite{jang2017categorical,maddison2017concrete}); gradients flow through the relation-specific projections, so interventions remain typed.

\textbf{Margin objective.} With the backbone frozen and the targets restricted to the instances being explained, we minimize
\begin{align}
\mathcal{L} &= \underbrace{\mathbb{E}_{v}\!\left[ \max(0,\, M_v(\G \odot m) + \kappa) \right]}_{\text{flip term}} \nonumber\\
&+ \underbrace{\lambda_{sp}\tfrac{1}{E}\textstyle\sum_{r,e}(1 - m^r_e)}_{\text{minimality}}
+ \underbrace{\beta\tfrac{1}{E}\textstyle\sum_{r,e} m^r_e (1 - m^r_e)}_{\text{binary pressure}},
\label{eq:cf}
\end{align}
replacing the soft-probability objective used by prior work: the hinge exerts no gradient on already-flipped instances, concentrating optimization on those near the boundary.

\textbf{Discrete verification and backward pruning.} The relaxed masks are \emph{not} the explanation. We (1)~order every edge by its learned keep-score and greedily delete in batches, evaluating the \emph{discrete} model after each batch and keeping the deletion point that maximizes validity minus $\lambda_{sp}$-weighted cost; (2)~backward-prune: restore deleted edges in descending keep-score while the satisfied set is preserved. The reported masks are therefore verified on the discrete model: unlike a threshold applied once after optimization, the selection itself consults the discrete model at every batch, so the flip status of the reported intervention is a measured property of that intervention rather than an implication of the relaxation.

\subsection{Instance-Level Relation Deletion Search}
\label{sec:reltype}

For each explained instance $v$, the relation-level counterfactual is the minimum-cost relation set whose deletion flips $v$:
\begin{equation}
S_v^{*} = \arg\min_{S \subseteq \{1,\dots,K\}}\; C_v(S)
\quad \text{s.t.} \quad M_v(\G \setminus S) \le -\kappa,
\label{eq:search}
\end{equation}
with the lexicographic cost
\begin{equation}
C_v(S) = \Big(|S|,\; \tfrac{1}{|E_L(v)|}\textstyle\sum_{r \in S} |E_r \cap E_L(v)|,\; M_v(\G \setminus S)\Big),
\label{eq:cost}
\end{equation}
i.e., fewest relation types first, then the smallest fraction of the instance's receptive-field edges, then the strongest flip. Because $K$ is small, we enumerate all $2^K$ subsets in increasing size; one forward pass per subset evaluates every instance simultaneously, so the total cost is $2^{K+1}$ forwards ($0.1$\,s at $K{=}3$, $10.3$\,s at $K{=}10$). The procedure is deterministic and returns the exact minimum under the frozen predictor and the stated cost. Instances with no feasible subset are marked infeasible, and the fraction of feasible instances is reported as \emph{coverage}: every instance receives either the certified minimum or an explicit infeasibility report, never a force-attributed relation. For larger vocabularies the budgeted branch-and-bound search of Sec.~\ref{sec:results} provides the same output contract. Two properties follow from the instance-level formulation and are measured in Sec.~\ref{sec:results}: (i)~the output contract---per instance, the certified answer or an explicit infeasibility report, with coverage disclosed ($0.17$--$0.33$ on our benchmarks); (ii)~non-monotonicity---some instances flip under a subset $S$ but not under full removal, so the aggregate ``$70$\% of the full-removal effect'' criterion of prior work can both miss and misattribute instances.

\subsection{Hierarchical Relation-to-Edge Refinement}
\label{sec:hier}

For every feasible instance, the relation-level answer is refined into a \emph{certified} typed edge set \emph{inside} $S_v^{*}$: starting from $\G \setminus S_v^{*}$ (already flipped), we restore edges of $S_v^{*}$ one at a time in descending keep-importance order (gradient saliency of the original prediction), accepting a restoration iff $v$ stays flipped, and repeat full passes until a complete pass changes nothing---the returned set is then \emph{single-edge-restoration irreducible}. If the per-instance forward budget is exhausted first, the answer is reported as a verified flip with the irreducibility check incomplete; the two statuses are reported separately, and a budget-exhausted answer is not marked optimal. The certificate is single-edge-restoration irreducibility rather than global minimality; the gap to the true minimum is measured exactly on enumerable subgraphs (Appendix~\ref{app:edge-cert}). Each check runs on the \emph{exact local computation subgraph} of $v$---the backward ball of radius $L$ with renumbered nodes, which reproduces the full-graph logits of $v$ exactly---so a check costs a tiny forward instead of a full-graph pass. Instances without a feasible relation set keep the flat verified edge explanation of Sec.~\ref{sec:edge} as a fallback, so overall success is never below the flat method. The output is the hierarchy: \emph{``delete relation $S_v^{*}$; within it, these edges.''}

\subsection{Fidelity Metrics and Overall Pipeline}
\label{sec:pipeline}

\begin{figure}[t]
\centering
\includegraphics[width=0.82\textwidth]{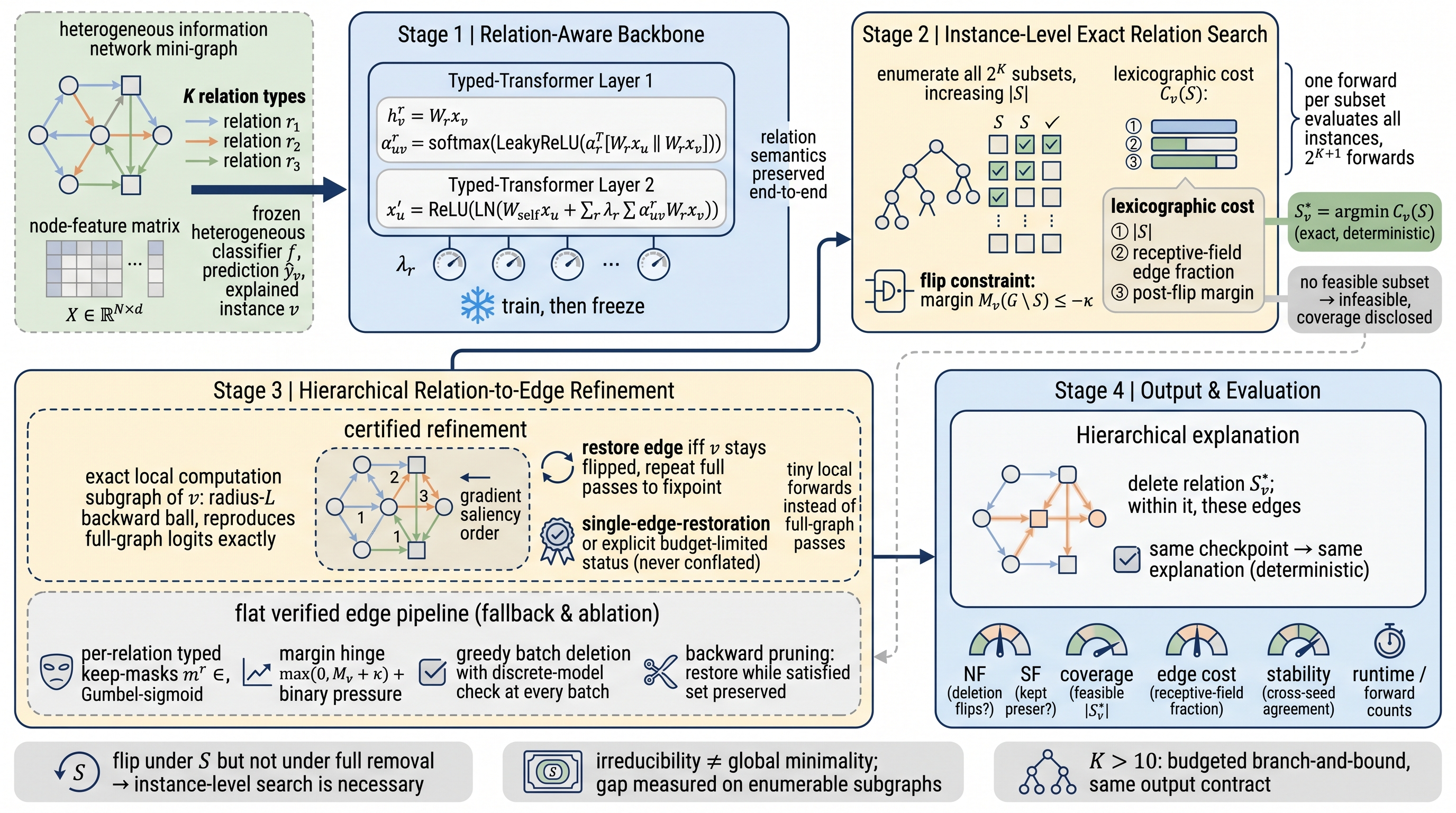}
\caption{Pipeline of \method{}. Stage 1 builds the relation-aware backbone and freezes it. Stage 2 solves the instance-level exact search, returning the certified minimum relation deletion set $S_v^{*}$ or an explicit infeasibility report. Stage 3 refines each feasible answer into a certified, single-edge-restoration-irreducible typed edge set (or an explicit budget-limited status); relation-infeasible instances keep the flat verified explanation as a fallback. Stage 4 evaluates NF/SF, coverage, cost, stability, and runtime.}
\label{fig:pipeline}
\end{figure}

Following the deletion/sufficiency evaluation convention of~\cite{cai2025pns} but without claiming causal probabilities, we report per instance and in aggregate:
\begin{equation}
\mathrm{NF}_v(S) = \mathbf{1}[f(\G \setminus S)_v \neq \hat y_v], \qquad
\mathrm{SF}_v(S) = \mathbf{1}[f(S)_v = \hat y_v],
\label{eq:nfsf}
\end{equation}
Necessity Fidelity (does the deletion flip?) and Sufficiency Fidelity (does the kept set alone preserve the prediction?). Aggregates are instance averages; we additionally report coverage, relation cost $|S_v^{*}|$, edge cost, the post-intervention margin, cross-seed stability, and runtime, and we never present $\mathrm{NF} \times \mathrm{SF}$ as a causal joint probability. Fig.~\ref{fig:pipeline} depicts the pipeline: \textbf{Stage 1} builds the relation-aware backbone (Sec.~\ref{sec:backbone}); \textbf{Stage 2} freezes it and solves the instance-level relation search of Eq.~\eqref{eq:search}, recording $S_v^{*}$ and feasibility; \textbf{Stage 3} refines each feasible answer into a certified, single-edge-restoration-irreducible typed edge set (Sec.~\ref{sec:hier}), with infeasible instances keeping the flat verified edge explanation; \textbf{Stage 4} evaluates both granularities with NF/SF, coverage, cost, margin, stability, and runtime.

\section{Experiments}
\label{sec:experiments}

We organize the study around six questions. \textbf{RQ1 (main comparison)}: on one frozen typed backbone, does the hierarchical \method{} improve over edge-level baselines on validity, cost, and stability? \textbf{RQ2 (mechanism)}: is the relation phase the source of the advantage, and what does discrete verification add? \textbf{RQ3 (output contract)}: does the instance-level search return the minimum valid relation deletion, with feasibility reported? \textbf{RQ4 (generality)}: does the advantage transfer across backbones and datasets? \textbf{RQ5 (determinism and robustness)}: how deterministic, stable, and perturbation-robust are the explanations, and what does verification cost? \textbf{RQ6 (scalability)}: how do exact and approximate search trade quality against cost as $K$ grows?

\subsection{Experimental Setup}
\label{sec:protocol}

\textbf{Datasets.} \label{sec:datasets} \emph{ACM (HAN version)}: a paper network whose relations are meta-path-derived---relation $0$: PAP (co-authorship), relation $1$: PTP (shared keywords)---$1{,}903$-dim bag-of-words features, venue labels. \emph{ogbn-mag (subsample)}~\cite{hu2020open}: $30{,}000$ papers of the top-$20$ venues, relations PAP and citation. \emph{Cora-derived}: citation plus a common-neighbor relation. \emph{ogbn-arXiv (subsample)}~\cite{hu2020open}: $30{,}000$ papers of the top-$20$ subject areas, relations direct citation and co-citation. \emph{DBLP}: an author network (co-authorship plus a shared-term meta-path relation; $4{,}057$ authors, $4$ areas). \emph{Synthetic SCM}: a five-mechanism generator (causal, spurious with environment shift, redundant, noise, and a synergistic variant). All four primary real benchmarks have $K=2$ relation types; a four-relation ogbn-mag variant and the synthetic $K$-scan probe larger vocabularies. Dataset statistics are summarized in Appendix~\ref{app:datasets}.

\textbf{Metrics.} We report backbone accuracy; \emph{counterfactual success rate} (CSR), the fraction of explained instances whose prediction flips, equal to mean NF, with any non-intervention counted as failure; \emph{edge cost}, the fraction of the instance's receptive-field edges deleted, under labeled aggregation conventions; NF/SF of Eq.~\eqref{eq:nfsf} separately; \emph{coverage}; cross-seed \emph{stability} (relation-set agreement and edge-set Jaccard); and \emph{runtime}. Synthetic studies add two separate hit rates: model-dependence recovery and data-causal recovery. Full definitions are in Appendix~\ref{app:metrics}.

\textbf{Baselines and protocol.} One typed backbone is trained per seed and frozen; \emph{every} explainer consumes the same checkpoint, the same explained instances (the correctly-classified test nodes), the same operation (edge deletion), and the same budget; objectives are restricted to the explained instances (``targeted''). Baselines: GNNExplainer-hetero, RACE-per-edge, PNS-hetero, CF\textsuperscript{2}-hetero, and the NSEG~\cite{cai2025pns}, InduCE~\cite{verma2024induce}, and C2Explainer~\cite{ma2025c2explainer} objectives, all operating on the typed graph with per-relation masks---the typed counterparts of the published methods, isolating the explanation objective from the architecture. We additionally compare the \emph{flat} verified pipeline (Sec.~\ref{sec:edge} without the relation phase) against the same baselines to isolate the hierarchical contribution (RQ2). HAN~\cite{wang2019heterogeneous}, HGT~\cite{hu2020heterogeneous}, and R-GCN~\cite{schlichtkrull2018rgcn} serve as the explained models in the cross-backbone study. Intervention semantics are fixed across every evaluation (deletion zeroes masked edge entries before aggregation; the attention softmax is not renormalized; node features and the residual self-projection are never modified). \textbf{Statistics.} Ten seeds on ACM/Cora/ogbn-mag (five for cross-backbone and additional-dataset studies); every comparison is a paired test over the same checkpoints and target instances; the primary comparisons are fixed in advance; $p$-values are exact two-sided sign-flip permutation tests (minimum attainable $p=0.002$ with ten seeds, $0.0625$ with five), supplemented by paired bootstrap 95\% CIs and paired Cohen's $d$; Holm correction is applied across the full family of $12$ primary tests. Full hyperparameters and settings are in Appendix~\ref{app:settings}.

\subsection{Results}
\label{sec:results}

\textbf{RQ1: main comparison.} Table~\ref{tab:main} reports the same-frozen-backbone, targeted comparison over ten seeds. At the edge level the flat explainers tie: RACE-per-edge and PNS-hetero are statistically identical on all three datasets ($p\ge0.38$), and the multi-term CF\textsuperscript{2}-hetero is at most $+0.1$~pp higher while deleting substantially more edges. The \emph{hierarchical} method breaks the tie: against the strongest baseline (CF\textsuperscript{2}-hetero) it improves CSR by $+0.8$/$+2.7$/$+1.4$~pp ($p=0.002$; Holm-corrected ${\le}0.062$ across the full test family) while deleting $-1.9$/$-1.0$/$-2.7$~pp \emph{fewer} edges ($p=0.002$), and its CSR gains over RACE-per-edge and PNS-hetero are significant on all three datasets ($p=0.002$). Prior methods trade success rate against deletion cost; the hierarchy improves both at once. Stability follows the same direction: pairwise cross-seed relation-set agreement is $0.779{\pm}0.058$/$0.641{\pm}0.096$/$0.894{\pm}0.020$ (ACM/Cora/mag), and the certified edge-set Jaccard is $0.436{\pm}0.041$/$0.483{\pm}0.072$/$0.654{\pm}0.032$---the separable datasets show the highest relation-set agreement; on Cora the relation sets vary more across seeds, an identifiability effect analyzed under RQ3 and Sec.~\ref{sec:scm}.

\begin{table}[t]
\centering
\caption{Same-frozen-typed-backbone, instance-targeted comparison (10 seeds, mean$\pm$std across seeds; CSR = success rate, cost = mean local deleted fraction over all targets). Paired sign-flip tests of the hierarchical method against each baseline; $\dagger$: $p=0.002$ before Holm correction (${\le}0.062$ after). \textbf{Bold}: best value in each column; \underline{underlined}: second best.}
\label{tab:main}
\resizebox{\textwidth}{!}{%
\small
\setlength{\tabcolsep}{4pt}
\begin{tabular}{llcccccc}
\toprule
& & \multicolumn{2}{c}{ACM} & \multicolumn{2}{c}{Cora} & \multicolumn{2}{c}{ogbn-mag} \\
& Method & CSR$\uparrow$ & cost$\downarrow$ & CSR$\uparrow$ & cost$\downarrow$ & CSR$\uparrow$ & cost$\downarrow$ \\
\midrule
\multirow{3}{*}{flat} & RACE-per-edge & \underline{0.216$\pm$0.072} & \textbf{0.159$\pm$0.062} & 0.452$\pm$0.111 & \underline{0.305$\pm$0.031} & \underline{0.168$\pm$0.008} & \underline{0.378$\pm$0.011} \\
& PNS-hetero & \underline{0.216$\pm$0.072} & \textbf{0.159$\pm$0.063} & 0.450$\pm$0.110 & \underline{0.306$\pm$0.031} & \underline{0.168$\pm$0.008} & \textbf{0.378$\pm$0.010} \\
& CF\textsuperscript{2}-hetero & \underline{0.216$\pm$0.072} & 0.195$\pm$0.088 & \underline{0.458$\pm$0.110} & 0.342$\pm$0.024 & \underline{0.168$\pm$0.008} & 0.434$\pm$0.013 \\
\midrule
\rowcolor{gray!12}\multicolumn{2}{l}{\textbf{RACE-hier}} & \textbf{0.224$\pm$0.077}$^{\dagger}$ & \underline{0.176$\pm$0.067}$^{\dagger}$ & \textbf{0.484$\pm$0.122}$^{\dagger}$ & 0.332$\pm$0.024$^{\dagger}$ & \textbf{0.182$\pm$0.009}$^{\dagger}$ & 0.407$\pm$0.011$^{\dagger}$ \\
\multicolumn{2}{l}{\hspace{1em}vs.\ CF\textsuperscript{2}-hetero} & $+0.8$~pp & $-1.9$~pp & $+2.7$~pp & $-1.0$~pp & $+1.4$~pp & $-2.7$~pp \\
\bottomrule
\end{tabular}}
\end{table}

\textbf{Comprehensive comparison.} Table~\ref{tab:zoo} compares all same-task counterfactual methods on the same frozen checkpoints with per-method RNG control. RACE-hier attains the highest CSR on every dataset ($0.224$/$0.484$/$0.182$), ahead of the strongest baseline by $+0.8$/$+2.6$/$+1.4$~pp. NSEG's necessity-sufficiency objective essentially never flips (CSR ${\le}0.01$), confirming that flip success is not what it optimizes; INDUCE is unstable at this scale (std up to $0.44$); C2 pays the highest deletion cost among the counterfactual objectives on ACM and Cora. The factual, gradient, and random diagnostics are reported separately in Appendix~\ref{app:diagnostics}: factual objectives do not optimize flips by construction, and the gradient attributions use an instance-blind global budget, so they are not same-task comparisons and are excluded from win claims. The comparison also isolates a property that only the exact search offers: the soft baselines' CSR varies substantially across runs that differ only in random ordering (CF\textsuperscript{2}: $0.216$ vs.\ $0.276$ on ACM, $0.452$ vs.\ $0.687$ on Cora), while \method{}'s relation-level answer is deterministic given the frozen backbone. On the nodes where both CF\textsuperscript{2} and RACE-hier flip, the hierarchical edge set is cheaper in $66\%$/$44\%$/$65\%$ of cases (ACM/Cora/ogbn-mag), and RACE-hier flips $109$/$85$/$394$ more nodes than CF\textsuperscript{2} while CF\textsuperscript{2} flips none that RACE-hier misses (Pareto frontier in Fig.~\ref{fig:pareto}, Appendix~\ref{app:diagnostics}).

\begin{table}[t]
\centering
\caption{Same-task comprehensive comparison under one frozen checkpoint family (10 seeds, mean$\pm$std). Cost annotations are restricted to the CSR columns, because a method that rarely flips can trivially attain low cost; the paper's cost claim is the paired comparison against CF\textsuperscript{2}-hetero (Table~\ref{tab:main}). RACE-hier additionally certifies every reported flip, provides relation-type attribution, and reports infeasibility per instance. \textbf{Bold}: best value; \underline{underlined}: second best.}
\label{tab:zoo}
\resizebox{\textwidth}{!}{%
\small
\setlength{\tabcolsep}{3pt}
\begin{tabular}{llcccccc}
\toprule
& & \multicolumn{2}{c}{ACM} & \multicolumn{2}{c}{Cora} & \multicolumn{2}{c}{ogbn-mag} \\
& Method & CSR$\uparrow$ & cost$\downarrow$ & CSR$\uparrow$ & cost$\downarrow$ & CSR$\uparrow$ & cost$\downarrow$ \\
\midrule
\multicolumn{8}{@{}l}{\textit{counterfactual}}\\
& CF-GNNExplainer & \underline{0.216$\pm$0.072} & 0.159$\pm$0.063 & 0.452$\pm$0.110 & 0.305$\pm$0.031 & \underline{0.168$\pm$0.008} & 0.379$\pm$0.011 \\
& RC & 0.148$\pm$0.049 & 0.145$\pm$0.059 & 0.400$\pm$0.100 & 0.245$\pm$0.013 & 0.088$\pm$0.007 & 0.314$\pm$0.012 \\
& NSEG & 0.000$\pm$0.001 & 0.139$\pm$0.041 & 0.001$\pm$0.001 & 0.224$\pm$0.021 & 0.002$\pm$0.001 & 0.192$\pm$0.014 \\
& CF\textsuperscript{2} & \underline{0.216$\pm$0.072} & 0.195$\pm$0.088 & \underline{0.458$\pm$0.110} & 0.342$\pm$0.024 & \underline{0.168$\pm$0.008} & 0.434$\pm$0.013 \\
& C2 & 0.190$\pm$0.069 & 0.406$\pm$0.023 & 0.448$\pm$0.113 & 0.518$\pm$0.024 & 0.123$\pm$0.007 & 0.275$\pm$0.013 \\
& INDUCE & 0.099$\pm$0.091 & 0.449$\pm$0.438 & 0.195$\pm$0.052 & 0.673$\pm$0.348 & 0.083$\pm$0.069 & 0.519$\pm$0.438 \\
\multicolumn{8}{@{}l}{\textit{search}}\\
& MEG & 0.072$\pm$0.017 & 0.342$\pm$0.004 & 0.115$\pm$0.030 & 0.342$\pm$0.010 & 0.061$\pm$0.006 & 0.348$\pm$0.007 \\
\midrule
\rowcolor{gray!12}\multicolumn{2}{l}{\textbf{RACE-hier (ours)}} & \textbf{0.224$\pm$0.077} & 0.176$\pm$0.067 & \textbf{0.484$\pm$0.122} & 0.332$\pm$0.024 & \textbf{0.182$\pm$0.009} & 0.407$\pm$0.011 \\
\bottomrule
\end{tabular}}
\end{table}

\textbf{RQ2: the relation phase is the mechanism.} Two controlled comparisons isolate where the advantage comes from. \emph{Flat verified pipeline.} Removing only the relation phase (greedy saliency deletion plus the same single-edge scan and budget, no relation search) leaves CSR at $0.174$/$0.246$/$0.165$ on ACM/Cora/ogbn-mag---below the strongest baseline and far below the hierarchy---at a lower edge cost (Table~\ref{tab:switch}); adding the relation phase converts this deficit into the significant CSR gains of Table~\ref{tab:main}. \emph{Relation-only} explanations (no edge refinement) reach only the coverage bound ($0.17$--$0.34$, Table~\ref{tab:instance}); the hierarchy strictly dominates both ends. \emph{Where the success comes from.} Table~\ref{tab:decomp} decomposes the ten-seed success rate: on ACM the relation phase contributes $18.2$ of the $22.4$~pp CSR at a mean cost of $0.105$---half the strongest baseline's cost ($0.195$); on ogbn-mag it delivers $16.7$ of $18.2$~pp at $0.273$ vs.\ $0.434$; on Cora it still delivers $33.7$ of $48.4$~pp at $0.311$ vs.\ $0.342$. \emph{Relation switch.} Table~\ref{tab:switch} runs the two pipelines under identical machinery and budgets---the only difference is the relation phase. The switch attributes $+5.0$/$+23.8$/$+1.7$~pp of CSR to the relation phase on ACM/Cora/ogbn-mag (all $p=0.002$) and $+5.2$/$+5.7$~pp on ogbn-arXiv/DBLP ($p=0.0625$, five seeds). On the nodes where both flip, the flat path deletes fewer edges (e.g., ACM $0.033$ vs.\ $0.107$), so the relation phase buys the significant CSR gain and the relation-type attribution at a higher per-explanation edge cost, while the pipeline-level cost advantage over CF\textsuperscript{2}-hetero is the separate, all-targets claim of Table~\ref{tab:main}. \emph{Verification.} Every reported flip is certified on the discrete model; the scan costs $1.5$--$6$~min per seed against $2$--$5$\,s for the soft-mask baselines---an inference-only cost that buys a per-instance certificate, attained for $35$--$83\%$ of the successful explanations (the remainder are reported as verified with the irreducibility check budget-limited). Budget curves and component ablations are in Appendix~\ref{app:mechanism}.

\begin{table}[t]
\centering
\caption{Relation switch (same machinery, scan budget, and target nodes; the only difference is the relation phase). CSR and feasible-success cost, mean$\pm$std over seeds; $p$: exact paired sign-flip. The relation phase buys a significant CSR gain at a higher per-explanation edge cost; the pipeline-level cost advantage over CF\textsuperscript{2}-hetero is the separate claim of Table~\ref{tab:main}.}
\label{tab:switch}
\small
\setlength{\tabcolsep}{3pt}
\begin{tabular}{lccccc}
\toprule
Dataset & hier CSR$\uparrow$ & flat CSR & $p$ & hier cost$\downarrow$ & flat cost$\downarrow$ \\
\midrule
ACM & \textbf{0.224$\pm$0.077} & 0.174$\pm$0.046 & 0.002 & 0.105 & \textbf{0.033} \\
Cora & \textbf{0.484$\pm$0.122} & 0.246$\pm$0.071 & 0.002 & 0.311 & \textbf{0.080} \\
ogbn-mag & \textbf{0.182$\pm$0.009} & 0.165$\pm$0.009 & 0.002 & 0.273 & \textbf{0.220} \\
ogbn-arXiv & \textbf{0.283$\pm$0.005} & 0.230$\pm$0.003 & 0.0625 & 0.287 & \textbf{0.139} \\
DBLP & \textbf{0.161$\pm$0.035} & 0.104$\pm$0.013 & 0.0625 & 0.157 & \textbf{0.030} \\
\bottomrule
\end{tabular}
\end{table}

\textbf{RQ3: exactness and the output contract.} Table~\ref{tab:instance} reports the instance-level relation search. Coverage is $0.18$ (ACM), $0.34$ (Cora), and $0.17$ (ogbn-mag): for most correctly-classified instances \emph{no} relation deletion flips the prediction, and \method{} says so per instance instead of force-attributing a relation. Where a deletion exists, it is cheap in relation count ($1.07$--$1.14$ relations on average) and strong (mean post-flip margin $-0.20$ to $-0.46$), with sufficiency fidelity $0.86$--$0.96$. The attribution is dataset-consistent: PAP on ACM (flip $0.150$ vs.\ $0.033$), citation on ogbn-mag ($0.130$ vs.\ $0.032$), and common-neighbor on Cora ($0.208$ vs.\ $0.130$, a consistent winner across all ten seeds); it also varies \emph{by class}, recovering per-class relation semantics that a dataset-level rate cannot (Appendix~\ref{app:contract}). Non-monotonicity is measurable: on ACM, coverage ($0.18$) \emph{exceeds} the full-removal flip rate ($0.14$)---instances flip under a subset but not under full deletion, evidence that dataset-level flip-rate searches cannot substitute for the instance-level problem. Four archetypes recur across the explained nodes (single-relation flip, two-relation flip, explicit refusal, flat-fallback success; Appendix~\ref{app:contract}), and bucketing by prediction margin shows the hierarchical CSR is highest exactly where predictions are closest to the decision boundary.

\begin{table}[t]
\centering
\caption{Instance-level relation search (10 seeds, mean$\pm$std). Coverage = fraction of correctly-classified test instances with any feasible relation deletion.}
\label{tab:instance}
\small
\setlength{\tabcolsep}{3pt}
\begin{tabular}{lccc}
\toprule
& ACM & Cora & ogbn-mag \\
\midrule
coverage $\uparrow$ & 0.182$\pm$0.048 & 0.337$\pm$0.096 & 0.167$\pm$0.009 \\
relation cost $\downarrow$ & 1.07$\pm$0.03 & 1.14$\pm$0.02 & 1.07$\pm$0.01 \\
edge cost $\downarrow$ & 0.574$\pm$0.013 & 0.605$\pm$0.044 & 0.838$\pm$0.014 \\
margin & $-0.263{\pm}0.033$ & $-0.459{\pm}0.042$ & $-0.197{\pm}0.015$ \\
SF $\uparrow$ & 0.918$\pm$0.031 & 0.858$\pm$0.028 & 0.959$\pm$0.009 \\
attribution & PAP (10/10) & common-nbr (10/10) & citation (10/10) \\
\bottomrule
\end{tabular}
\end{table}

\textbf{RQ4: generality across backbones and datasets.} Table~\ref{tab:arxiv} reports ogbn-arXiv under four frozen backbones (five seeds each): the hierarchical CSR advantage over the strongest baseline reproduces on $4/4$ backbones ($+0.9$ to $+2.6$~pp, all $p=0.062$), and the cost advantage over CF\textsuperscript{2}-hetero on $4/4$ ($-4.1$ to $-7.1$~pp); R-GCN, whose simpler aggregation makes more predictions relation-deletable, yields the highest CSR ($0.490$). The backbone study extends to ACM, Cora-derived, and ogbn-mag (three backbones each, five seeds; Table~\ref{tab:crossbb}, Appendix~\ref{app:generality}): the hierarchical method beats CF\textsuperscript{2}-hetero on CSR in all $9/9$ cells ($+1.0$ to $+8.1$~pp, all $p=0.062$). DBLP reproduces the CSR advantage---$0.161$ vs.\ $0.145$ for CF\textsuperscript{2}-hetero ($+1.6$~pp, five seeds, $p=0.0625$). \emph{Natural multi-relation vocabulary.} A four-relation variant of ogbn-mag (PAP, citation, shared-field-of-study, shared-institution) raises coverage from $0.17$ to $0.27$, and the hierarchy still beats CF\textsuperscript{2}-hetero ($0.322$ vs.\ $0.299$, $+2.3$~pp, ten seeds, $p=0.002$) and the flat switch ($+9.3$~pp, $p=0.002$), with the attribution staying on citation (5/5 seeds). Taken with Table~\ref{tab:main}, the mechanism is dataset- and backbone-agnostic, and the relation-level contract is not an artifact of $K{=}2$ vocabularies.

\begin{table}[t]
\centering
\caption{ogbn-arXiv: hierarchical vs.\ strongest baseline per backbone (5 seeds; all CSR/cost comparisons $p=0.062$, the minimum attainable with five seeds). \method{}-bb is the relation-aware backbone of Sec.~\ref{sec:backbone}. \textbf{Bold}: better value in each pairwise comparison.}
\label{tab:arxiv}
\small
\setlength{\tabcolsep}{4pt}
\begin{tabular}{lcccc}
\toprule
& \multicolumn{2}{c}{CSR$\uparrow$} & \multicolumn{2}{c}{cost$\downarrow$} \\
\cmidrule(lr){2-3}\cmidrule(lr){4-5}
Backbone & RACE-hier & CF\textsuperscript{2} & RACE-hier & CF\textsuperscript{2} \\
\midrule
\method{}-bb & \textbf{0.283} & 0.267 & \textbf{0.287} & 0.385 \\
HAN & \textbf{0.255} & 0.245 & \textbf{0.284} & 0.379 \\
HGT & \textbf{0.288} & 0.275 & \textbf{0.286} & 0.430 \\
R-GCN & \textbf{0.490} & 0.464 & \textbf{0.315} & 0.510 \\
\bottomrule
\end{tabular}
\end{table}

\textbf{RQ5: determinism and robustness.} Given the frozen backbone, the relation-level answer is exact and deterministic; across seeds the relation sets agree at $0.89$/$0.62$/$0.89$ (ACM/Cora/mag), and under 5\% random edge add/remove the attributed relation sets remain near-identical ($S_v^{*}$ agreement $0.975{\pm}0.016$/$0.930{\pm}0.023$/$0.965{\pm}0.011$). Sensitivity ablations (Table~\ref{tab:abl}, Appendix~\ref{app:robustness}) confirm the reported results are not artifacts of the margin threshold or the cost definition: $\kappa{=}0.05$ lowers coverage by $-2.3$/$-1.0$/$-3.3$~pp (the expected effect of a stricter flip threshold), and switching the cost criterion leaves coverage identical on all three datasets.

\textbf{RQ6 and the Synthetic SCM: scalability and identifiability.}
\label{sec:scm}
A synthetic five-mechanism generator (CF-GNNExplainer-style~\cite{lucic2022cfgnnexplainer}; causal, spurious with environment shift, redundant, noise, and a synergistic variant; generation details in Appendix~\ref{app:scm}) separates \emph{model-dependence} recovery (the returned set equals the relation the trained model relies on) from \emph{data-causal} recovery (it equals the generating relation). In the benchmark regime ($K\le10$), exact search certifies the minimum-cost answer at negligible wall-clock cost ($0.1$\,s at $K{=}3$ to $7$\,s at $K{=}10$, faster than the effect-ordered greedy/beam heuristics for $K\le5$), and the double hit-rate protocol reports model-dependence equal to data-causal recovery---the regularized model relies on the generating relation. The heuristics attain higher data-causal hit only because the lexicographic tie-break prefers cheaper noise relations over the causal one when both flip; exact search's guarantee is the certified minimum-cost answer. When the mechanism is not learnable (synergistic variant, test accuracy $0.25\approx$ chance), all data-causal hits collapse to chance while model-dependence sits at $0.20$: the double protocol attributes the collapse to model unlearnability rather than explainer error, which a single hit rate conflates. At $K{=}20$, where $2^{20}$ enumeration is impractical, the budgeted branch-and-bound search (a first-order Lipschitz bound, capped at $500$ local forwards per instance) is the only approximation that improves over greedy/beam ($0.16$ vs.\ $0.06$ data-causal hit). The full results are in Table~\ref{tab:scm} (Appendix~\ref{app:scm}); the same certification discipline applies at the edge level, where the scan equals the true minimum in $85\%$ of enumerable cases and is off by one edge in the rest (Appendix~\ref{app:edge-cert}).

\section{Conclusion}
\label{sec:conclusion}

We presented \method{}, which explains a frozen heterogeneous GNN at two granularities that prior work conflates: an \emph{instance-level} minimal relation-deletion set found by exact, deterministic search, and a \emph{verified}, single-edge-restoration-irreducible typed edge set inside the attributed relations. Under a strict same-frozen-typed-backbone, instance-targeted protocol, the hierarchical explainer significantly improves success rate while deleting fewer edges than the strongest baseline on ACM, Cora-derived, and ogbn-mag; the advantage reproduces across four backbones on ogbn-arXiv, on DBLP, and on a natural four-relation ogbn-mag; and across the same-task counterfactual baselines, \method{} attains the highest success rate on every dataset. Two properties make the answers dependable in practice: determinism---the same checkpoint yields the same explanation, with cross-seed relation-set agreement up to $0.89$ and $0.93$--$0.98$ agreement under $5\%$ edge perturbation---and the per-instance output contract---every reported flip is certified on the discrete model, and every infeasible instance is reported as such rather than force-attributed. Future work extends the intervention unit from relation types to meta-paths, grounds relation-level fidelity in explicit structural causal models, scales the certified search past $K{=}10$ with provable approximation bounds, and evaluates hierarchical explanations with human users.

\subsection*{AI use statement}

In this work, we used generative AI tools to aid and polish the writing of the paper. We have not used generative AI tools for any task requiring disclosure---generating synthetic data, developing theoretical models or conceptual frameworks, formulating mathematical claims or assisting their proofs, proposing or refining hypotheses, designing or providing feedback on research methodology or experiments, implementing methods, translation, cleaning or reformatting datasets, supporting qualitative or thematic data analysis, or interpreting results; these tasks are not applicable to this work. All research ideas, methods, experiments, analyses, figures, tables, and claims were produced and verified by the authors. We have reviewed all AI-assisted work and take responsibility for the final content of this work, including text, claims, and artifacts produced with the aid of generative AI.

\subsection*{Reproducibility statement}
\label{sec:repro}

All datasets used in this work are public benchmarks (ACM, ogbn-mag, ogbn-arXiv, Cora, DBLP) or synthetic generators described in Sec.~\ref{sec:scm}; the subsampling and preprocessing steps are specified in Sec.~\ref{sec:datasets} and Appendix~\ref{app:settings}. The complete experimental protocol---hyperparameters, intervention semantics, and statistical procedures---is given in Sec.~\ref{sec:protocol} and Appendix~\ref{app:settings}. An anonymized implementation of \method{} (backbone training, exact relation search, hierarchical refinement, and the evaluation harness) is available at \url{https://anonymous.4open.science/r/race_official-68A1}.

\bibliographystyle{iclr2027_conference}
\bibliography{refs}

\appendix
\raggedbottom

\section{Dataset Statistics}
\label{app:datasets}

Table~\ref{tab:datasets} summarizes the datasets used in the experiments.

\begin{table}[H]
\centering
\caption{Datasets used in the experiments.}
\label{tab:datasets}
\footnotesize
\setlength{\tabcolsep}{3pt}
\begin{tabular}{lcccp{2.6cm}}
\toprule
Dataset & \#Nodes & \#Classes & Features & Relation types \\
\midrule
ACM (HAN) & 12,499 & 14 & 1,903-d BoW & PAP, PTP \\
ogbn-mag (sub) & 30,000 & 20 & 128-d emb. & PAP, citation \\
ogbn-mag (sub, 4-rel) & 30,000 & 20 & 128-d emb. & PAP, citation, PFP, PAIP \\
Cora-derived & 2,708 & 7 & 1,433-d BoW & citation, common-neighbor \\
ogbn-arXiv (sub) & 30,000 & 20 & 128-d emb. & citation, co-citation \\
DBLP & 4,057 & 4 & 334-d BoW & co-authorship, shared-term \\
Synthetic SCM & 400 & 4 & 16-d noise & causal, spurious, redundant, noise $\times (K{-}3)$ \\
\bottomrule
\end{tabular}
\end{table}

\section{Extended Experimental Settings}
\label{app:settings}

\textbf{Backbone.} Hidden dim $32$, $2$ layers, dropout $0.5$, weight decay $5\times10^{-4}$, early stopping. \textbf{Explainer.} $150$ mask steps (Adam, lr $0.1$, $\lambda_{sp}=0.05$, Gumbel $\tau=1.0$), then discrete verification ($40$--$80$ deletion batches) and up to three backward-pruning sweeps. \textbf{Relation search.} Exact enumeration, $\kappa=0$; the margin constraint of Eq.~\eqref{eq:search} is applied uniformly in the relation search and in the discrete edge-level checks (ties count against the original class). \textbf{Intervention semantics.} Deletion zeroes the masked edge entries before aggregation; the attention softmax is \emph{not} renormalized over the surviving edges (it is computed on the intact edge set); node features and the residual self-projection are never modified; edge sets are stored symmetrically so each direction is an independent maskable message entry, and all costs count these directed message entries. Every check runs the frozen model in eval mode with the original predicted class held fixed. \textbf{Verification cost.} The restoration scan costs $1.5$--$6$~min per seed against $2$--$5$\,s for the soft-mask baselines, with the cost concentrating in the per-node restoration passes (relation enumeration $0.1$--$0.7$\,s, one saliency pass $0.01$\,s). \textbf{Statistics.} Aggregates over seeds use the per-seed (macro) mean unless a table caption says otherwise; hierarchical fallback records are imputed at the per-seed CF\textsuperscript{2} mean cost in the all-targets cost convention of Table~\ref{tab:main}.

\section{Extended Metric Definitions}
\label{app:metrics}

\textbf{Backbone accuracy.} Test accuracy of $f$ on the intact graph---reported separately from explanation quality (the explained model, not the explainer). \textbf{Counterfactual success rate (CSR).} The fraction of explained instances whose prediction flips under the intervention; equals mean NF. A method that returns no intervention (infeasible, or budget exhaustion without a verified flip) counts as a failure, never as a zero-cost success. \textbf{Edge cost.} The fraction of the instance's receptive-field edges deleted (reported per instance and averaged); relation cost is $|S_v^{*}|$. Cost is reported under labeled conventions---over all targets (failures assigned the full-cost penalty $1.0$), over each method's successful subset, and over the subset on which all compared methods succeed---and no aggregate mixes them without a label. \textbf{NF / SF.} Necessity and sufficiency fidelity of Eq.~\eqref{eq:nfsf}, reported separately. \textbf{Coverage.} The fraction of explained instances with a feasible relation-level deletion (for the relation-level and hierarchical methods). \textbf{Stability.} Cross-seed agreement of the relation set (identity agreement) and Jaccard of the deleted edge sets; for baselines, Jaccard of their masks across seeds. \textbf{Runtime.} Wall-clock per seed, plus forward-pass counts for the search procedures. \textbf{Hit rates (synthetic only).} Model-dependence recovery (the returned set equals the relation the trained model actually relies on) and data-causal recovery (the returned set equals the generating relation)---reported separately, since they need not agree.

\section{Diagnostic Comparison and Pareto Frontier}
\label{app:diagnostics}

\begin{figure}[H]
\centering
\includegraphics[width=0.95\textwidth]{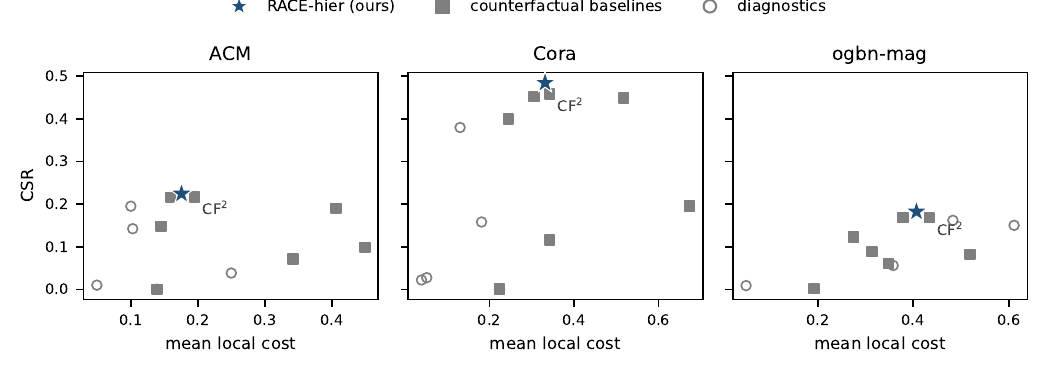}
\caption{Success rate vs.\ mean local deletion cost across the compared methods (10 seeds; same-task counterfactual methods in filled markers, diagnostics in open markers). RACE-hier (star) is the highest-CSR point on all three datasets, Pareto-dominating the strongest baseline CF\textsuperscript{2} (annotated).}
\label{fig:pareto}
\end{figure}

\begin{figure}[H]
\centering
\includegraphics[width=0.7\textwidth]{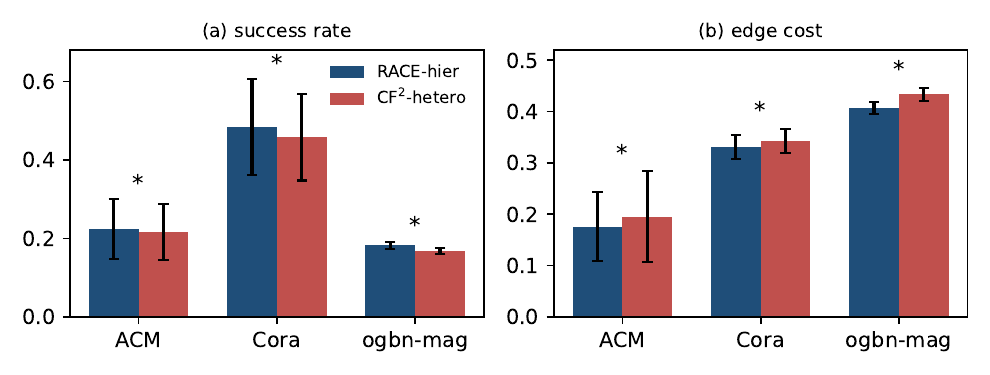}
\caption{Hierarchical vs.\ the strongest baseline (CF\textsuperscript{2}-hetero): (a) counterfactual success rate and (b) mean local edge cost (10 seeds, mean $\pm$ 1 std; $*$: paired sign-flip $p=0.002$, Holm-corrected ${\le}0.062$). The hierarchy improves both metrics at once on every dataset.}
\label{fig:csrcost}
\end{figure}

Table~\ref{tab:zoodiag} reports the factual, gradient, and random diagnostics on the same checkpoints. Factual objectives (GNNExplainer, PGExplainer) optimize prediction preservation, so near-zero CSR is by construction; the gradient attributions delete a single instance-blind global top-$5\%$ edge budget and are not per-instance interventions; Random is a sanity floor. These rows are not same-task baselines and are excluded from win claims. Figure~\ref{fig:pareto} shows the success-rate-vs-cost frontier across all compared methods, and Fig.~\ref{fig:csrcost} contrasts the hierarchical method with the strongest baseline on CSR and cost directly.

\begin{table}[H]
\centering
\caption{Diagnostic comparison (same checkpoints, 10 seeds, mean$\pm$std). Factual objectives optimize prediction preservation, so near-zero CSR is by construction; the gradient attributions delete a single instance-blind global top-$5\%$ edge budget and are not per-instance interventions; Random is a sanity floor. These rows are not same-task baselines and are excluded from win claims.}
\label{tab:zoodiag}
\resizebox{\textwidth}{!}{%
\small
\setlength{\tabcolsep}{3pt}
\begin{tabular}{llcccccc}
\toprule
& & \multicolumn{2}{c}{ACM} & \multicolumn{2}{c}{Cora} & \multicolumn{2}{c}{ogbn-mag} \\
& Method & CSR$\uparrow$ & cost$\downarrow$ & CSR$\uparrow$ & cost$\downarrow$ & CSR$\uparrow$ & cost$\downarrow$ \\
\midrule
\multicolumn{8}{@{}l}{\textit{factual}}\\
& GNNExplainer & 0.000$\pm$0.001 & 0.139$\pm$0.041 & 0.001$\pm$0.001 & 0.224$\pm$0.021 & 0.002$\pm$0.001 & 0.192$\pm$0.014 \\
& PGExplainer & 0.038$\pm$0.058 & 0.250$\pm$0.354 & 0.022$\pm$0.048 & 0.039$\pm$0.080 & 0.056$\pm$0.062 & 0.358$\pm$0.412 \\
\multicolumn{8}{@{}l}{\textit{gradient (global top-5\% edge budget, instance-blind)}}\\
& Saliency & 0.142$\pm$0.026 & 0.103$\pm$0.006 & 0.158$\pm$0.037 & 0.181$\pm$0.030 & 0.150$\pm$0.007 & 0.611$\pm$0.012 \\
& GradCAM & 0.195$\pm$0.053 & 0.100$\pm$0.005 & 0.379$\pm$0.080 & 0.131$\pm$0.015 & 0.162$\pm$0.008 & 0.483$\pm$0.012 \\
\multicolumn{8}{@{}l}{\textit{random}}\\
& Random & 0.010$\pm$0.003 & 0.050$\pm$0.001 & 0.028$\pm$0.010 & 0.052$\pm$0.004 & 0.009$\pm$0.002 & 0.050$\pm$0.003 \\
\bottomrule
\end{tabular}}
\end{table}

\section{Mechanism: CSR Decomposition, Budget Curves, and Component Ablation}
\label{app:mechanism}

\emph{CSR decomposition.} Table~\ref{tab:decomp} decomposes the ten-seed success rate into the relation-phase answers and the flat fallback (per-seed contributions, so the two phases sum to the total by construction). The flat fallback only extends coverage where no relation deletion exists, so the hierarchy is not a wrapper around the flat explainer but the mechanism that produces the double advantage.

\begin{table}[H]
\centering
\caption{Where the success rate comes from (10 seeds): relation-phase answers vs.\ the flat fallback. The relation phase delivers most of the CSR at roughly half the strongest baseline's cost.}
\label{tab:decomp}
\small
\setlength{\tabcolsep}{3pt}
\begin{tabular}{lccc}
\toprule
& ACM & Cora & ogbn-mag \\
\midrule
relation-phase answers (certified, $n$) & 2546 & 1106 & 4185 \\
\quad CSR contribution (pp) & 18.2$\pm$4.8 & 33.7$\pm$9.6 & 16.7$\pm$0.9 \\
\quad mean edge cost (fraction) & 0.105 & 0.311 & 0.273 \\
fallback CSR (pp) & 4.2$\pm$2.9 & 14.7$\pm$2.9 & 1.5$\pm$0.3 \\
\midrule
\textbf{total CSR} & \textbf{0.224} & \textbf{0.484} & \textbf{0.182} \\
CF\textsuperscript{2} mean cost (ref.) & 0.195 & 0.342 & 0.434 \\
\bottomrule
\end{tabular}
\end{table}

\emph{Component ablation.} Table~\ref{tab:b14} isolates each component on the same checkpoints. Removing the flat fallback drops CSR by exactly the fallback's contribution ($-4.2$/$-14.7$/$-1.5$~pp, matching Table~\ref{tab:decomp}); the scan budget trades cost against compute along the budget curves (coarser budgets are faster but coarser; the default budget $128$ is the working point used throughout). The ``full'' row reproduces the numbers of Table~\ref{tab:main}, as the pipeline's core is deterministic given the frozen checkpoint; the rows are derived from the per-node scan traces, so no separate runs were needed.

\begin{table}[H]
\centering
\caption{Component ablation of the hierarchical pipeline (10 seeds, mean$\pm$std; CSR over all targets, cost = mean feasible edge cost at the given scan budget). full = saliency prior + CF\textsuperscript{2} fallback + single-edge scan (budget 128).}
\label{tab:b14}
\resizebox{\textwidth}{!}{%
\small
\setlength{\tabcolsep}{3pt}
\begin{tabular}{lllllll}
\toprule
Variant & \multicolumn{2}{c}{ACM} & \multicolumn{2}{c}{Cora} & \multicolumn{2}{c}{ogbn-mag} \\
 & CSR$\uparrow$ & cost$\downarrow$ & CSR$\uparrow$ & cost$\downarrow$ & CSR$\uparrow$ & cost$\downarrow$ \\
\midrule
full (budget 128) & 0.224$\pm$0.077 & 0.107 & 0.484$\pm$0.122 & 0.313 & 0.182$\pm$0.009 & 0.273 \\
$-$ fallback & 0.182$\pm$0.048 & 0.107 & 0.337$\pm$0.096 & 0.313 & 0.167$\pm$0.009 & 0.273 \\
budget 64 & 0.182 & 0.176 & 0.322 & 0.390 & 0.167 & 0.273 \\
budget 32 & 0.182 & 0.279 & 0.322 & 0.458 & 0.167 & 0.287 \\
\bottomrule
\end{tabular}}
\end{table}

\begin{figure}[H]
\centering
\includegraphics[width=0.7\textwidth]{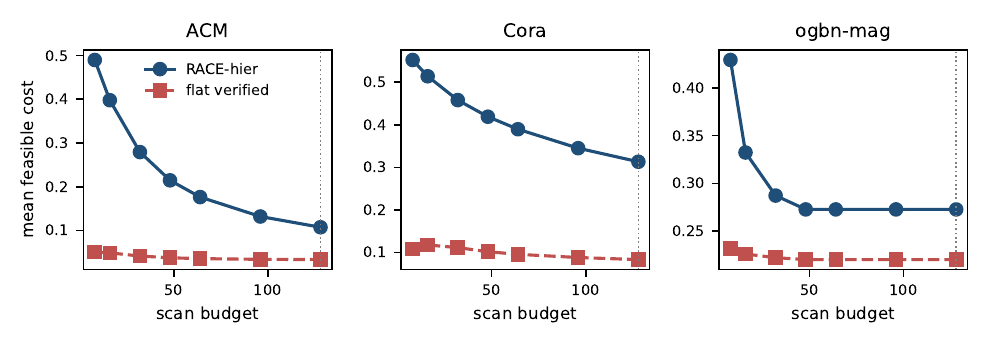}
\caption{Budget curves (10 seeds): mean feasible edge cost vs.\ the per-node scan budget for the hierarchical (solid) and flat-verified (dashed) pipelines, reconstructed from the per-node scan traces. The hierarchy's cost decreases monotonically with budget at unchanged CSR; the flat variant saturates after $\le16$ forwards. The default working point is budget $128$.}
\label{fig:budget}
\end{figure}

\emph{Budget curves.} Reconstructing the per-node scan traces at forward budgets $\{8,\dots,128\}$ (Fig.~\ref{fig:budget}) shows the hierarchical cost decreasing monotonically with budget (ACM $0.490{\to}0.107$; Cora $0.552{\to}0.313$) at unchanged CSR, while the flat variant saturates after $\le16$ forwards; the curves give the common working point at any budget.

\section{Output Contract: Attribution Detail, Archetypes, and Explanation Structure}
\label{app:contract}

\begin{figure}[H]
\centering
\includegraphics[width=0.7\textwidth]{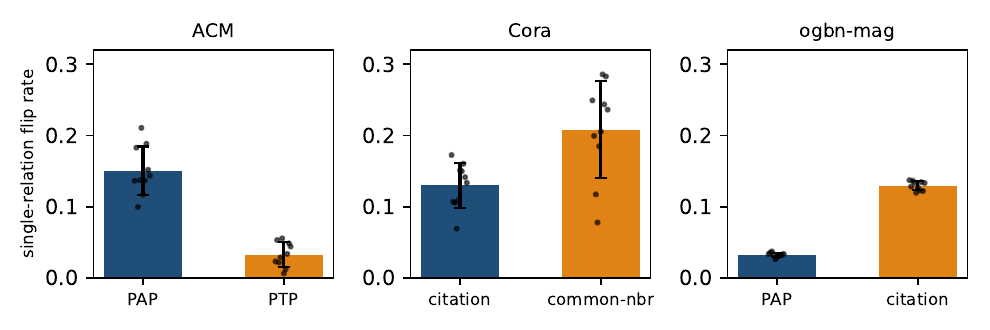}
\caption{Single-relation flip rates (10 seeds, mean $\pm$ 1 std, dots: per-seed values): ACM (separable: co-authorship PAP dominates), Cora (common-neighbor wins with moderate separation), ogbn-mag (citation dominates).}
\label{fig:fliprates}
\end{figure}

\emph{Per-class attribution.} On ogbn-mag, classes 1/2/6/11/13/16/17/18 are citation-dominated ($68$--$100\%$ of relation-attributed nodes) while classes 0/3/4/14/19 lean on co-authorship ($31$--$60\%$); on ACM, co-authorship dominates every venue class ($62$--$86\%$) with the keyword relation contributing $6$--$30\%$. The instance-level search thus recovers per-class relation semantics that a dataset-level rate cannot. Figure~\ref{fig:fliprates} reports the single-relation flip rates behind these attributions.

\begin{figure}[H]
\centering
\includegraphics[width=0.7\textwidth]{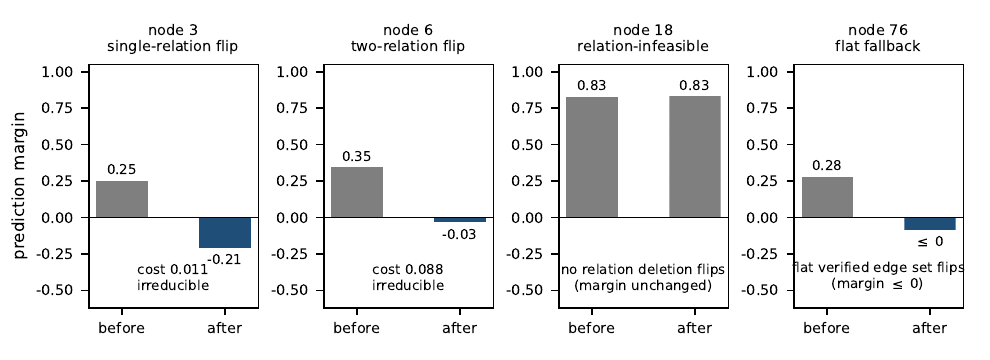}
\caption{Case studies (ACM, seed 0): the four archetypes of the output contract. (a) a single-relation flip; (b) a two-relation flip; (c) an explicit refusal---no relation deletion flips the prediction, reported as infeasible; (d) a relation-infeasible node that the flat verified edge set flips instead. Bars show the prediction margin before/after the intervention (hatched: no relation-level change exists); edge costs are the deleted fraction of the node's receptive field.}
\label{fig:casestudy}
\end{figure}

\emph{Archetypes.} Four archetypes recur across the explained nodes (ACM, seed 0; Fig.~\ref{fig:casestudy}): a single-relation flip (node 3: delete co-authorship, $1/87$ local edges, margin $-0.21$, irreducible certificate), a two-relation flip (node 6: co-authorship plus keyword, $8/91$ local edges, margin $-0.032$, irreducible), an explicit refusal (node 18: no relation deletion flips, margin $0.83$), and a flat-fallback success (node 76). The same pipeline returns a human-readable relation answer where one exists, and says so where it does not.

\begin{figure}[H]
\centering
\includegraphics[width=0.7\textwidth]{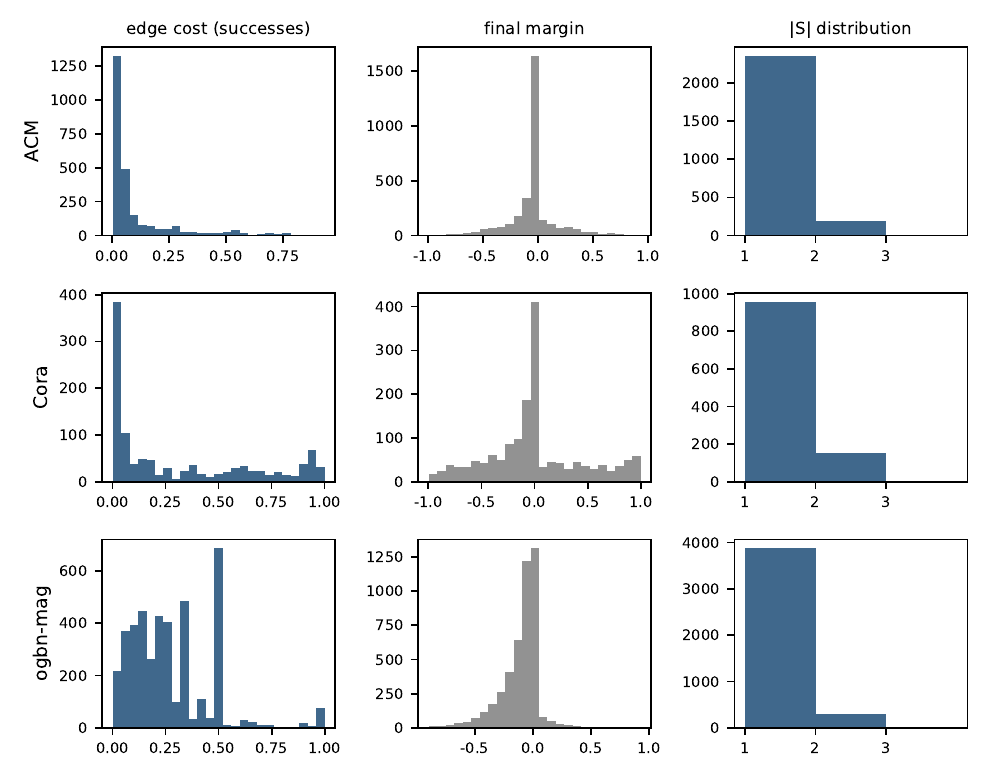}
\caption{Structure of the explanations (10 seeds): the certified edge sets are cheap (mean local cost $0.11$--$0.31$); a single relation suffices for $86$--$93\%$ of feasible instances; and the final margins separate flipped from refused instances.}
\label{fig:dist}
\end{figure}

\emph{Explanation structure.} Figure~\ref{fig:dist} summarizes the structure of the explanations: the certified edge sets are cheap (mean local cost $0.11$--$0.31$); a single relation suffices for $86$--$93\%$ of feasible instances; and the final margins separate flipped from refused instances. \emph{Margin buckets.} Bucketing the explained nodes by their prediction margin (10 deciles, 10 seeds, pooled), the hierarchical CSR is highest in the lowest-margin decile and falls monotonically across deciles on all three datasets (ACM $0.693$ to $0.015$; Cora $0.885$ to $0.177$; ogbn-mag $0.396$ to $0.030$): the method delivers relation-level counterfactuals exactly where predictions are closest to the decision boundary---the instances for which an explanation is most consequential.

\section{Generality: Cross-Backbone Study}
\label{app:generality}

Table~\ref{tab:crossbb} extends the backbone study of Table~\ref{tab:arxiv} to ACM, Cora-derived, and ogbn-mag (three backbones each, five seeds): the hierarchical method beats CF\textsuperscript{2}-hetero on CSR in all $9/9$ backbone--dataset cells ($+1.0$ to $+8.1$~pp, all $p=0.062$, the minimum attainable with five seeds). The paired cost advantage is established in the all-targets comparison of Table~\ref{tab:main} and on ogbn-arXiv (Table~\ref{tab:arxiv}).

\begin{table}[H]
\centering
\caption{Cross-backbone study on ACM / Cora-derived / ogbn-mag (5 seeds; hierarchical vs.\ CF\textsuperscript{2}-hetero, CSR $=$ success rate; all CSR differences $p=0.062$, the minimum attainable with five seeds). The CSR advantage holds in all $9/9$ backbone--dataset cells; the paired cost advantage is established in the all-targets comparison of Table~\ref{tab:main} and on ogbn-arXiv (Table~\ref{tab:arxiv}). \textbf{Bold}: better value in each pairwise comparison.}
\label{tab:crossbb}
\footnotesize
\setlength{\tabcolsep}{4pt}
\begin{tabular}{llccc}
\toprule
Backbone & Method & ACM CSR$\uparrow$ & Cora CSR$\uparrow$ & ogbn-mag CSR$\uparrow$ \\
\midrule
\multirow{2}{*}{HAN} & RACE-hier & \textbf{0.255} & \textbf{0.314} & \textbf{0.182} \\
& CF\textsuperscript{2}-hetero & 0.244 & 0.298 & 0.164 \\
\midrule
\multirow{2}{*}{HGT} & RACE-hier & \textbf{0.384} & \textbf{0.611} & \textbf{0.205} \\
& CF\textsuperscript{2}-hetero & 0.355 & 0.581 & 0.185 \\
\midrule
\multirow{2}{*}{R-GCN} & RACE-hier & \textbf{0.785} & \textbf{0.939} & \textbf{0.318} \\
& CF\textsuperscript{2}-hetero & 0.771 & 0.858 & 0.292 \\
\bottomrule
\end{tabular}
\end{table}

\section{Robustness and Sensitivity Ablations}
\label{app:robustness}

Table~\ref{tab:abl} reports the $\kappa$-sensitivity and cost-criterion ablations plus relation-set robustness under 5\% random edge add/remove: $\kappa{=}0.05$ lowers coverage by $-2.3$/$-1.0$/$-3.3$~pp (the expected effect of a stricter flip threshold, ACM/Cora/mag); switching the cost criterion leaves coverage identical on all three datasets ($0.1818$/$0.3370$/$0.1674$); and the attributed relation sets remain near-identical under perturbation.

\begin{table}[H]
\centering
\caption{Sensitivity ablations (ACM/Cora/mag, 10 seeds): margin threshold and cost criterion, plus relation-set robustness under 5\% random edge add/remove.}
\label{tab:abl}
\small
\setlength{\tabcolsep}{4pt}
\begin{tabular}{@{}>{\raggedright\arraybackslash}p{2.2cm}>{\raggedright\arraybackslash}p{4.2cm}>{\raggedright\arraybackslash}p{6.5cm}@{}}
\toprule
Ablation & Setting & Result (ACM / Cora / ogbn-mag) \\
\midrule
\textit{margin threshold $\kappa$} & $\kappa{=}0.05$ vs.\ $\kappa{=}0$; coverage & $0.159{\pm}0.044$ vs.\ $0.182$;\quad $0.327{\pm}0.095$ vs.\ $0.337$;\quad $0.135{\pm}0.009$ vs.\ $0.167$ \\
\addlinespace[3pt]
\textit{cost criterion} & type / edge / lexicographic; coverage & identical on all datasets ($0.1818$ / $0.3370$ / $0.1674$) \\
\addlinespace[3pt]
\textit{edge perturbation} & 5\% random add/remove; $S_v^{*}$ agreement & $0.975{\pm}0.016$ / $0.930{\pm}0.023$ / $0.965{\pm}0.011$ \\
\bottomrule
\end{tabular}
\end{table}

\section{Edge-Level Certification on Enumerable Subgraphs}
\label{app:edge-cert}

On enumerable computation subgraphs (synthetic SCM, 1-layer backbone so radius-1 computation subgraphs are enumerable; 120 instances) we certify the restoration scan against full edge enumeration (Table~\ref{tab:c1}): the certified heuristic equals the true minimum deletion count in $102/120$ cases and is off by one edge in the five other feasible cases; non-monotonicity (a flip that disappears when one edge is added back) occurs in $60/120$ instances.

\begin{table}[H]
\centering
\caption{Exact edge-level certification on enumerable subgraphs (120 instances). The certified heuristic equals the true minimum deletion count in $102/120$ cases and is off by one edge in the five other feasible cases.}
\label{tab:c1}
\small
\setlength{\tabcolsep}{4pt}
\begin{tabular}{lcc}
\toprule
Quantity & Value & Note \\
\midrule
scan $=$ true minimum & $102/120$ ($85\%$) & scan vs.\ full enum. \\
off by exactly one edge & $5$ & only nonzero gaps \\
non-monotonic instances & $60/120$ ($50\%$) & flip under $S$, not $S\cup\{e\}$ \\
\bottomrule
\end{tabular}
\end{table}

\section{Synthetic SCM: Additional Detail}
\label{app:scm}

\emph{Generation.} Features are iid noise; labels are determined by the mean of the \emph{causal}-relation neighbors' features projected onto class patterns; a \emph{spurious} relation is homophilic in an attribute correlated with the label in the train environment but independent in the test environment; a \emph{redundant} relation duplicates the causal signal; the rest are Erd\H{o}s--R\'enyi noise; a \emph{synergistic} variant splits the label into two parts recoverable only from two relations jointly. For each configuration and ten data seeds we train the frozen backbone and report the two hit rates of Sec.~\ref{sec:scm}. Table~\ref{tab:scm} reports the full $K$-scan: exact search attains $0.529$/$0.292$/$0.151$ data-causal hit at $K{=}3/5/10$ while the effect-ordered heuristics attain $0.634$/$0.440$/$0.286$.

\begin{table}[H]
\centering
\caption{Synthetic SCM (10 data seeds; hit = data-causal recovery; model-dependence recovery coincides with it at $K{\le}10$ and is $0.089$ at $K{=}20$ / $0.201$ synergistic; exact at $K\le10$, greedy/beam everywhere, bnb at $K{=}20$). Exact search certifies the minimum-cost answer and is the fastest search in the benchmark regime ($K\le10$); greedy/beam are effect-ordered heuristics, and at $K{=}20$ the budgeted branch-and-bound (bnb) is the only approximation that improves over them.}
\label{tab:scm}
\small
\setlength{\tabcolsep}{3pt}
\begin{tabular}{>{\raggedright\arraybackslash}p{2.5cm}ccccc}
\toprule
Config & exact & greedy & beam & bnb & time (ex./gr.) \\
\midrule
standard $K{=}3$ & 0.529 & 0.634 & 0.633 & --- & 0.07\,s / 1.0\,s \\
standard $K{=}5$ & 0.292 & 0.440 & 0.438 & --- & 0.23\,s / 2.5\,s \\
standard $K{=}10$ & 0.151 & 0.286 & 0.285 & --- & 7.0\,s / 5.6\,s \\
standard $K{=}20$ & --- & 0.064 & 0.064 & 0.158 & --- / 33\,s \\
synergistic $K{=}5$ & 0.018 & 0.018 & 0.021 & --- & 0.17\,s / 1.1\,s \\
\bottomrule
\end{tabular}
\end{table}

\emph{Irreducibility vs.\ optimality.} A scripted adversarial instance instantiates the irreducibility-vs-optimality gap: valid deletion sets $\{a,b,c\}\supset\{a,b\}\supset\{c\}$ coexist, the restoration scan (saliency order $a>b>c$) returns the irreducible $\{a,b\}$ (cost 2), and $\{c\}$ (cost 1) is the true minimum---the certificate characterizes the returned set, and the gap is measured exactly where enumeration is feasible (Appendix~\ref{app:edge-cert}).

\emph{Synergistic variant.} When the mechanism is not learnable (test accuracy $0.25\approx$ chance; the failure decomposes as compounded per-part errors), all hits collapse to chance on the data-causal target while model-dependence sits at $0.20$; the double hit-rate evaluation makes the cause visible---the drop reflects model unlearnability, not explainer error---which a single hit rate conflates.

\begin{figure}[H]
\centering
\includegraphics[width=0.8\textwidth]{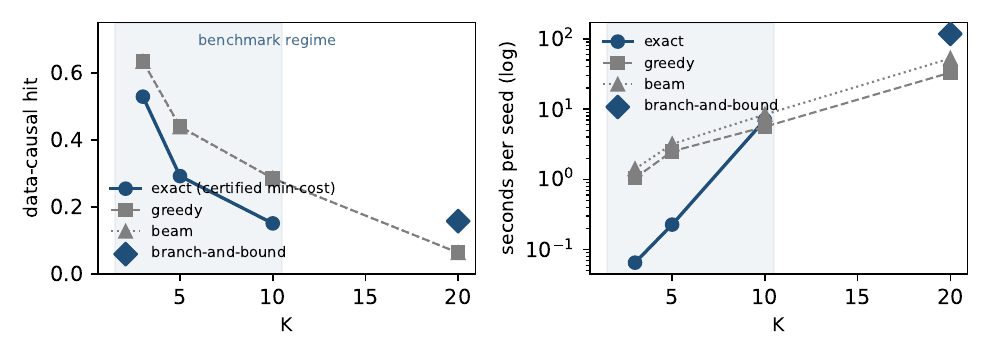}
\caption{Relation search as the vocabulary grows (synthetic SCM, 10 data seeds, mean $\pm$ 1 std; shaded: benchmark regime $K{\le}10$). (a)~Data-causal hit rate: exact search certifies the minimum-cost answer at every feasible $K$; the effect-ordered greedy/beam heuristics rank the causal relation higher because the lexicographic cost prefers cheaper noise relations; at $K{=}20$ the budgeted branch-and-bound (bnb) is the only approximation that improves over greedy/beam. (b)~Wall-clock per seed: exact search is the fastest in the benchmark regime.}
\label{fig:kscaling}
\end{figure}

\emph{Scaling.} Figure~\ref{fig:kscaling} reports the hit rates and wall-clock cost of all search procedures as the relation vocabulary grows; exact search is the fastest search in the benchmark regime ($K{\le}10$) and certifies the minimum-cost answer at every feasible $K$, while at $K{=}20$ reliance drifts onto noise relations (test accuracy $0.33$) and the budgeted branch-and-bound is the only approximation that improves over greedy/beam (companion implementation).

\end{document}